\documentclass[conference]{IEEEtran}
\IEEEoverridecommandlockouts
\usepackage{cite}
\usepackage{amsmath,amssymb,amsfonts}
\usepackage{algorithmic}
\usepackage{graphicx}
\usepackage{textcomp}
\usepackage{xcolor}

\def\BibTeX{{\rm B\kern-.05em{\sc i\kern-.025em b}\kern-.08em
    T\kern-.1667em\lower.7ex\hbox{E}\kern-.125emX}}
\begin{document}

\title{Value of Information-Enhanced Exploration \\in Bootstrapped DQN
\thanks{\hrule \smallskip \noindent Extended pre-print of an article appearing as a {\em full paper}  in the Proceedings of IJCNN-2025, the 2025 INNS International Joint Conference on Neural Networks. Please cite as: 
S. Plataniotis, C. Akasiadis and G. Chalkiadakis, "Value of Information-Enhanced Exploration in Bootstrapped DQN" 2025 International Joint Conference on Neural Networks (IJCNN), Rome, Italy, 2025, pp. 1-8, doi: 10.1109/IJCNN64981.2025.11229044.}
}

\author{\IEEEauthorblockN{Stergios Plataniotis}
\IEEEauthorblockA{\textit{Technical University of Crete} \\
Chania, Greece \\
splataniotis@tuc.gr}
\and
\IEEEauthorblockN{Charilaos Akasiadis}
\IEEEauthorblockA{\textit{Technical University of Crete \& NCSR ``Demokritos''} \\
Chania, Greece \\
cakasiadis@iit.demokritos.gr}
\and
\IEEEauthorblockN{Georgios Chalkiadakis}
\IEEEauthorblockA{\textit{Technical University of Crete} \\
Chania, Greece \\
gchalkiadakis@tuc.gr}
}

\maketitle

\begin{abstract} Efficient exploration in deep reinforcement learning remains a fundamental challenge, especially in environments characterized by high-dimensional states and sparse rewards. Traditional exploration strategies that rely on random local policy noise, such as $\epsilon$-greedy and Boltzmann exploration methods, often struggle to efficiently balance exploration and exploitation. In this paper, we integrate the notion of (expected) value of information (EVOI) within the well-known Bootstrapped DQN algorithmic framework, to enhance the algorithm's deep exploration ability. Specifically, we develop two novel algorithms that incorporate the expected gain from learning the value of information into  Bootstrapped DQN. Our methods use value of information estimates to measure the discrepancies of opinions among distinct network heads, and drive exploration towards areas with the most potential. We evaluate our algorithms with respect to performance and their ability to exploit inherent uncertainty arising from random network initialization. Our experiments in complex, sparse-reward Atari games demonstrate increased performance, all the while making better use of uncertainty, and, importantly, without introducing extra hyperparameters.
\end{abstract}

\begin{IEEEkeywords}
Machine learning, Value of information
\end{IEEEkeywords}

\section{Introduction}

Reinforcement Learning (RL) \cite{sutton2018reinforcement} is a rigorous paradigm that 
enables agents to learn how to make optimal decisions. The ultimate goal for an agent, is to 
derive an optimal policy (i.e. a mapping of states to the best corresponding actions) that maximizes the cumulative reward. 
RL is often paired with the notion of Markov 
Decision Processes (MDPs) \cite{sutton2018reinforcement}. 
In short, an MDP models the problem of 
decision-making and consists of a state space, an 
action space and transition 
and reward 
functions that encapsulate the environment's 
dynamics. 
The advent of deep neural networks 
has further enabled RL to deal 
efficiently with high-dimensional inputs---thus, giving 
rise to deep reinforcement learning (DRL)~\cite{arulkumaran2017deep, franccois2018introduction}. 
In the context of RL, efficient exploration involves constantly 
acquiring new information to improve the policy. 
This remains a significant challenge, especially 
in environments with sparse or 
delayed rewards, where the agent may need to 
take seemingly uninteresting actions so as 
to 
optimize sequential performance.

As an example, consider a simple scenario  
which essentially contains two paths for an agent:  the first 
leads to an immediate reward of $+1$ after one step, while the other leads to 
a reward of $+10$ after a sequence of 
rewardless states. A learning method  
acting greedily will discover the first path and 
exploit it indefinitely. However, a method that explores 
efficiently would avoid settling for the 
short-term reward and continue exploring in search of a better outcome. This can be 
very challenging if the ``sequence of rewardless states'' is long enough, and the problem could 
be deemed as a hard-exploration one.

Value-based methods aim to learn a function 
that maps the state-action pairs $(s,a)$ to estimated values of the expected cumulative reward for taking action $a$ at a state $s$. These values are termed as $Q$-values. 
One of the most well-known methods in RL is $Q$-learning~\cite{watkins1992q}. This algorithm 
maintains and updates a table of the $Q$-values.  
The agent can then simply look up the corresponding values for each action 
at its current state, and utilize 
this table in its action-selection procedure. 
Building on this foundation,~\cite{mnih2015human} developed a 
DRL variant of $Q$-learning to tackle the 
highly complex environment of Atari 2600 games 
by observing raw frames. 
The resulting algorithm, Deep Q-Networks (DQN), employs a deep neural network to replace the table, making this DRL method applicable to high dimensional domains.

Ensemble methods, such as Bootstrapped DQN \cite{osband2016deep}, involve multiple models 
that, in turn, provide multiple estimates of the $Q$-function. 
At the start of an episode, a random head 
is assigned, and its policy is followed 
until a terminal signal is received. 
The varied behavior of distinct models in Bootstrapped DQN arises from 
random initialization and bootstrapping of collected data and can be naturally used to estimate uncertainty in 
decisions~\cite{pathak2019self,da2020uncertainty}. 
This leads to uncertainty-aware methods which 
are superior to more naive dithering strategies, such as $\epsilon$-greedy~\cite{sutton2018reinforcement, hao2023exploration}. Henceforth, we refer to 
this algorithm as either Bootstrapped 
DQN or simply \textit{BootDQN}, using  both terms interchangeably.

Building on the strengths of ensemble methods, 
our work integrates a powerful concept from decision theory; the \emph{Expected Value of  
Information} (EVOI) \cite{dearden1998bayesian,dearden1999model,chalkiadakis2003coordination,teacy2012decentralized,russell2020artificial}. 
EVOI aims to quantify the information 
the algorithm would gain by 
choosing an action given a specific  state. 
This new data is expected to provide new insights and enhance the method's policy. 
In this paper, we adapt EVOI for use in a deep reinforcement learning context, 
using an ensemble of models. In 
our approach, EVOI estimates the expected 
benefit of exploratory actions based on the discrepancies 
among the value-function estimates. By guiding the 
agent to take actions that are expected to maximize 
information gain, we enhance the quality of exploration and improve learning performance.

We introduce two novel algorithms
which extend Bootstrapped DQN by incorporating uncertainty-aware mechanisms. By leveraging ensemble methods and the EVOI, we utilize the inherent diversity of a network ensemble to calculate the 
expected gain from learning the value of information 
for any state-action pair, enabling more principled and directed exploration. 
Our first method, \textit{BootDQN-Gain} calculates the expected gain of information from the point-of-view of the active head, 
while our second method, \textit{BootDQN-EVOI}, computes EVOI as the {\em aggregated} (expected) gain of information over all heads. 
We demonstrate the performance of our algorithms in several complex Atari 2600 games \cite{bellemare2013arcade}, showing their superiority over vanilla Bootstrapped DQN. Notably, \textit{BootDQN-EVOI} increases Bootstrapped DQN’s performance by 9.38\% in maximal human-normalized scores across all tested games, achieving this improvement without the need for extra hyperparameters. 
Moreover, our techniques lead to 
more diverse policies among individual models in most scenarios. 

In summary, in this paper we propose the use of the gain of information and EVOI concepts for the leveraging of the uncertainty inherent in the decision-making process of 
the members of ensemble DRL methods. 
Our approach enables the 
agent to benefit from the diverse 
perspectives of the ensemble, enhancing 
its exploration capabilities.
Our findings suggest potential new 
directions for developing more robust  
and effective exploration strategies in 
ensemble DRL.

\section{Background and Related Work}
\label{section:backround}

Bootstrapped DQN uses and updates 
a single deep neural network that 
branches into multiple output heads. 
Each of these heads is initialized 
with different weights and has its 
own target function to compute its 
own loss. Bootstrapping in training of the heads is also applied by sharing 
each stored transition tuple with 
other heads with a probability 
sampled from a pre-determined 
distribution. The diversity that 
sources from these applications 
drives deep exploration as the heads 
learn diverse aspects of the problem 
at hand. At the start of each 
episode, a head is selected 
uniformly at random and its policy 
is followed greedily until a 
terminal signal is received. The genuine increase in performance is 
observed during evaluation periods 
where collective knowledge can be  
used, either by averaging $Q$-values 
over heads or utilizing a voting 
mechanism \cite{brandt2016handbook}, to select actions. 
Despite its admittedly increased  
performance compared to DQN, 
\cite{lin2024the} demonstrates that 
Bootstrapped DQN and ensemble methods in general suffer from the 
\emph{curse of diversity} as each 
head's individual performance is subpar 
compared to DQN.

The idea of making use of the 
distinct heads of Bootstrapped 
DQN to harness uncertainty has been investigated in 
other works as well, although these 
approaches differ from our 
proposed technique. 
For example, \cite{nikolov2019information} adapts Information-Directed 
Sampling \cite{russo2014learning} 
to the context of RL, with \textit{DQN-IDS} being 
one variant of their approach. 
\textit{DQN-IDS} utilizes 
Bootstrapped DQN's multiple value-estimates to calculate the empirical mean and standard deviation over all heads. These metrics are then used 
to estimate the regret and 
information gain associated with taking a 
specific action. This 
is employed in the 
action-selection strategy, to opt for actions 
that balance minimizing regret and 
acquiring valuable information. The information 
gain function used by \textit{DQN-IDS}, is simply a 
non-linear transformation of the variance.

Then, Chen {\em et al.} in
\cite{chen2017ucbexplorationqensembles} define the information gain by leveraging the 
entropy of multiple $Q$-functions. Specifically, 
the Boltzmann distribution for 
each network head's output is calculated and these are aggregated to obtain an average 
distribution across all heads. The potential information gain for each head can then be 
computed by comparing the individual distributions to the average using KL-divergence \cite{kullback1951information}. 
The overall information gain of a given state is 
obtained by averaging these distances across all 
heads, and is taken into account as an extrinsic reward signal, 
augmenting the rewards received from the environment. 
This approach is more intrusive than simply 
guiding action-selection, since it interferes with 
rewards and consequently with loss computation and 
$Q$-value shaping. 
An
upper confidence bound (UCB)\footnote{Upper-confidence
bound (UCB) algorithms for multi-arm bandits maintain a UCB for each arm, {\em s.t.} its expected reward is smaller than its UCB with high probability.
At any given step, the arm with the highest
UCB is selected.} 
 variant
was also used \cite{audibert2009,auer2002finite}, choosing
actions that maximize the sum of the mean $Q$-values and their
standard deviation.

Instead of quantifying the gain 
of information as the variance of 
values or as the distance between 
distributions, we advocate the 
use of 
the
{\em Expected Value of Information (EVOI)}~\cite{chalkiadakis2003coordination, teacy2012decentralized} to select 
actions that 
are expected to provide
useful insights about the true 
optimal policy. 

Chalkiadakis and Boutilier in~\cite{chalkiadakis2003coordination} extend the work of \cite{dearden1999model} and utilize EVOI 
in cooperative multi-agent environments through a 
Bayesian model-based approach. Each agent 
maintains 
a belief that incorporates distributions that model both the environment dynamics and other agents. 
With this compact domain representation, an agent can 
sample several MDPs and solve them to 
obtain various value functions $Q_i(s,a)$. 
For any given state, the agent compares the values from   
each MDP to the average across all MDPs to 
estimate the expected improvement in decision 
quality from taking a specific action. 
This estimation is then integrated into the 
decision-making process, enhancing  
action-values to guide exploration.

Teacy {\em et al.}~\cite{teacy2012decentralized} reiterate the effectiveness 
of EVOI in Bayesian model-based RL, and introduce 
a convenient formula for EVOI that extends its application to a model-free approach in a multi-agent scenario. Instead of
maintaining densities over reward and transition functions, 
they directly model a distribution over possible 
$Q$-functions. 
From this, one 
can easily obtain statistics such as the expected $Q$-values, and use them to compute EVOI. 
Their model-free  
EVOI version
avoids the need for multiple MDPs, 
thereby 
reducing the 
computational overhead by 
directly focusing on the $Q$-values. 

\begin{figure}[t]
    \centering    \includegraphics[width=\columnwidth]{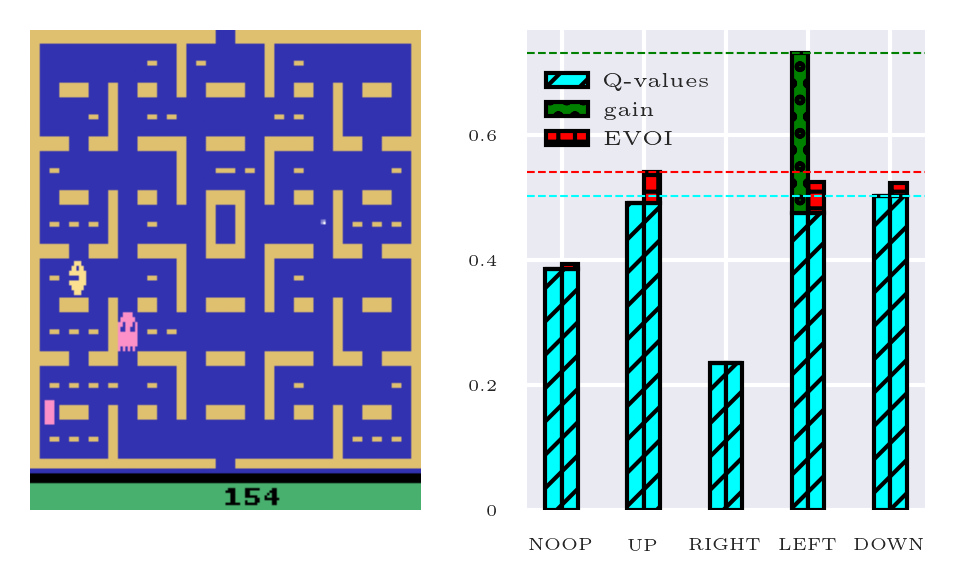}
    \caption{
    Example of the effect of different action-selection methods. 
    Left: a specific state of the game in Pacman. Right: a barplot that depicts for each action the \textit{BootDQN}-computed $Q$-values (blue part of each bar), augmented with either $gain$ (left bars, $gain$ contribution in green) or EVOI (right bars, EVOI contribution in red). 
    Horizontal dashed lines illustrate the 
    maximum values for each method, depicted in corresponding colors. 
    \textit{BootDQN}, in this 
    state, would choose action DOWN; \textit{BootDQN-Gain} would choose LEFT; and \textit{BootDQN-EVOI} would choose UP. This showcases the effect of our methods.}
    \label{fig:$EVOI$_showcase}
\end{figure}

By  
choosing actions with high EVOI, the agent performs meaningful 
exploration in more uncertain parts of the 
state-space. Despite 
its success
in small environments, 
this pre-deep RL era
approach
is not 
readily usable in
high-dimensional problems. To this 
end, we adjust this method in the 
context of Bootstrapped DQN by using 
the different network heads---instead 
of multiple MDPs---to calculate the 
expected gain of 
information relative to either a particular head (\textit{BootDQN-Gain}) or the collective view of all heads (\textit{BootDQN-EVOI}). Fig. 
\ref{fig:$EVOI$_showcase} provides 
intuitions on how these two methods 
operate.

\section{Our Approach}
\label{section:our_algorithms}

Bootstrapped DQN maintains and updates a group of $K$ bootstrap network heads, resulting in multiple estimates $Q_{k\in[1,K]}$ of the $Q$-function. 
Each one of them offers a unique 
perspective, while collectively providing 
a means to measure uncertainty at a given 
state.  
In our work in this paper, we put forward an approach that 
leverages these estimates in order to compute and utilize the value of 
information towards insightful and strategic action choices; and by so doing, to facilitate a novel, principled exploration method for 
ensemble-based DRL methods. 
This approach 
involves engaging the distinct 
heads to resolve the epistemic uncertainty as described by their respective $Q$-values---uncertainty that originates from the random initialization of each head's individual layers.

Before delving into the details of our contributions, we present the pseudocode 
for our implementation of Bootstrapped DQN in 
Algorithm~1. 
In line 1, 
we initialize a neural network 
with a common backbone that branches 
into $K$ network heads and approximates $K$ value-functions $Q_{k\in[1,K]}$. 
In line 2 an active head is selected uniformly 
at random. In line 3 we obtain the environment's initial state.

\begin{figure}[t]
\raggedright
\hrule
\textbf{Algorithm 1} BootDQN pseudocode \\ 
\hrule
\renewcommand{\algorithmicrequire}{\textbf{Input:}}
\centering
\begin{algorithmic}[1]
    \REQUIRE Number of heads $K$, timesteps $T$, network update frequency $N$, sharing parameter $p$, replay buffer $B$
    \STATE Initialize $K$ value-functions $Q_{k \in [1,K]}$ 
    \STATE $h \sim \mathcal{U}_{[1,K]}$ \COMMENT{Sample random head to follow}
    \STATE Obtain initial state $s_0$
    \FOR{$t = 0, 1, \ldots, T$}
        \STATE $a_t \leftarrow$ action\_selection($s_t, Q_{k \in [1,K]}, h$) \COMMENT{Greedy selection by default}
        \STATE Act on the environment with action $a_t$ and obtain $s_{t+1}, r_{t+1}, \textit{terminal}_{t+1}$
        \STATE $m_t \sim \text{Bernoulli}(p)$ \COMMENT{Sample a binary mask}
        \STATE Store $(s_t, a_t, s_{t+1}, r_{t+1}, \textit{terminal}_{t+1}, m_t)$ in $B$
        \IF{$t \bmod N = 0$}
            \STATE Sample experience batch $b \sim B$
            \STATE Compute $loss_h$ for each head $h$ w.r.t. $b$ \COMMENT{Only samples $b_i$ with $m^i_h=1$ contribute in loss}
            \STATE Backpropagate normalized loss $\frac{1}{K} \sum_h loss_h$
            \STATE Perform a gradient step
        \ENDIF
        \IF{$\textit{terminal}_{t+1}$}
            \STATE Reset the environment and obtain state $s_{t+1}$
            \STATE $h \sim \mathcal{U}_{[1,K]}$
        \ENDIF
        \STATE $s_t \leftarrow s_{t+1}$
    \ENDFOR
\end{algorithmic}
\hrule
\label{fig:BootDQN}
\end{figure}

Lines 5-19 constitute the core of the method, 
iterating at each timestep until reaching a predetermined number of steps. 
In line 5, we use an exploration 
strategy to select an action, 
defaulting to greedy selection according to the acting head. Lines 6-8 involve  
interacting with the environment using 
the selected action, receiving  
new information, and storing it in the replay buffer along with a sampled 
binary mask $m_t$. 
Value $m_{t,h}$ of the 
mask dictates 
whether or not head 
$h$ may utilize the 
corresponding experience 
tuple for training. 

Lines 10-13 are performed every $N$ steps 
to update the parameters of the neural network. 
In this context, a mini-batch of past experiences $b$ 
is sampled from the replay buffer and contains a number of tuples. 
Each sample $b_i=(s^i_t,a^i_t,s^i_{t+1},r^i_{t+1},m^i_{t}), i \in [1,|b|]$ 
contains information regarding the 
transition from timestep $t$ to $t+1$.  
$s^i_t,a^i_t,m^i_t$ are the state, 
action taken and binary mask at 
timestep $t$ while $s^i_{t+1}$ and $r^i_{t+1}$ are the 
subsequent state and reward received. 
For each $b_i$ and each head $h$, the target value is computed using the double $Q$-learning update introduced by \cite{van2016deep}: 
\[target^i_h = r^i_{t+1} + \gamma \cdot Q^-_h(s^i_{t+1}, \arg\,max_a Q_h(s^i_{t+1}, a))\]
where $\gamma$ is the discount factor that  ensures the cumulative reward is finite, 
and $Q^-$ represents the so-called target network, as 
proposed by \cite{mnih2015human}. 
The target network is used to improve stability 
in learning. Its parameters are synced with those of the main-policy network every $C$ 
steps, but otherwise remain constant. 
Once $target_h^i$ is computed, the loss for each 
head is calculated, typically using the 
mean squared error: 
\[loss_h = \frac{1}{\sum_i m^i_{t,h}}\sum_i m^i_{t,h} (Q_h(s^i_t,a^i_t) - target^i_h)^2\]
though more sophisticated loss functions, such 
as the Huber loss \cite{huber1964robust} can 
be employed. 
We highlight here the 
effect of the binary mask $m^i_{t,h}$ (sampled from a Bernoulli distribution) which 
negates the contribution of sample 
$b_i$ in head's total loss if $m^i_{t,h}=0$. 
The losses are 
then aggregated by averaging over all heads to stabilize 
learning, and the mean loss is 
backpropagated before the gradient step. 

Finally, in lines 16-17, if the newly 
encountered state is terminal, indicating that the episode ends, the simulator is reset, and a new head is sampled to follow in the 
next episode. 

We note that methods that use the standard deviation of the 
$Q$-values in action-selection~\cite{chen2017ucbexplorationqensembles, nikolov2019information, lee2021sunrise} are closely related to our approach. However, our methods consider action ranking to compute a more meaningful evaluation of potential information gain, respective to the current optimal policy. 

\subsection{Gain of Information}
Upon observation of the current state of the environment, we can 
obtain the outputs ($Q$-values) from all the neural network heads. This results in $K$ values for each action denoted as $Q_k(s,a)$. We can then compute the average $Q$-value for each action, 
across all $K$ heads, $\overline{Q}(s,a)=\frac{\sum_{k}Q_k(s,a)}{K}$.  
These aggregated values represent the most-recent estimated collective estimate of the value of each action. By comparing the distance between each ensemble member's values and the aggregated ones, we can quantify the expected information gain by taking a specific action.

Next, we  
sort the actions based on $\overline{Q}$ and store the best and second-best actions,  
denoted as $a^*_1$ and $a_2^*$, respectively. 
Notably, information-gain is calculated from a 
specific head's perspective as it evaluates collective knowledge to determine whether it has misjudged the value of an action. This data can be used for more informed action 
selection by prioritizing actions that we expect to better the existing policy. 

Specifically, with $\overline{Q}$, $a_1^*$ and $a_2^*$ available, and from the perspective of head $h$, we examine each available action 
$a$ along with its value $Q_h(s,a)$. 
If head $h$ has undervalued the best action ($Q_h(s,a_1^*)<\overline{Q}(s,a_2^*)$), 
we quantify $gain$  as the discrepancy between $h$'s evaluation of the best action and the average evaluation of the second-best action: \[gain_{h}(s,a_1^*)=\overline{Q}(s,a^{*}_2) - Q_h(s,a_1^*)\]
On the other hand, if head $h$ has overvalued an action $a \neq a_1^*$ such that $Q_h(s,a) > \overline{Q}(s,a^*_1)$, we set $gain$ as the discrepancy between $h$'s evaluation of action $a \neq a_1^*$ and the average evaluation of the best action: \[gain_h(s,a)=Q_h(s,a) - \overline{Q}(s,a_1^*)\]
Otherwise we assume that $gain_h(s,a)=0$. Equation~\ref{equation:gain} formulates this behavior:

\begin{equation}
gain_h(s,a)=
\left\{
\begin{array}{ll}
    \max(\overline{Q}(s,a^*_2)-Q_h(s,a),0) \textnormal{,} & a=a^*_1\\
    \max(Q_h(s,a)-\overline{Q}(s,a^*_1),0) \textnormal{,} & a \neq a^*_1
\end{array}
\right.
\label{equation:gain}
\end{equation}
Hereafter, we refer 
to this computed vector simply as $gain$. 
Computing $gain$ can be performed expeditiously in parallel using tensor operations instead of iterating over all actions.

Intuitively, $gain_h$ will be high 
for actions that $h$ has misjudged according to group knowledge and low 
for actions that have $Q$-values comparable to the average ones. 
We underscore that  $\overline{Q}(s,a_2^*)$ is used when $a = a^*_1$, and 
$\overline{Q}(s,a_1^*)$ 
is used when $a \neq a^*_1$, in order to 
mitigate the effects 
of outliers.

Since in Bootstrapped DQN discrepancies originate from the random 
initialization of the network, heads are bound to disagree more in rare states than in states that the agent is familiar with. 
This diversity in policies is the fundamental strength 
of this algorithm and holds  
even when heads share information 
with a high probability\footnote{Notably, during testing, we 
observed a build-up of diversity 
during the initial stage of training}. 
Therefore, we need methods that harness 
this uncertainty. 

To this end, we attempt to make each head aware of what information 
is missing, through the utilization of aggregated knowledge, in order to make more informed decisions. We argue that 
this can be accomplished with the use of $gain$. Specifically, instead of 
choosing the action with the highest $Q_h$-value (according to 
head/selector $h$), we propose the selection of actions that maximize 
both $Q_h(s,a)$ (individual knowledge) and $gain_h(s,a)$ (collective knowledge from $h$'s perspective), as follows: \[arg\,max_{a}\,Q_h(s,a)+ gain_h(s,a)\] 
As such, we value exploiting the current optimal policy and acquiring new information equally.  
We term \textit{BootDQN} with this action selection procedure as \textit{BootDQN-Gain}.

\subsection{Expected Value of Information (EVOI)}

Building upon the previous approach,  instead of computing the $gain$ from the 
point-of-view of each head, we can compute the EVOI as the aggregated gain of 
information. We set EVOI to be equal to the average $gain$ over all heads: 
\begin{equation}
    EVOI(s,a) = \frac{\sum_{k \in K}gain_k(s,a)}{K}
    \label{equation:vpi}
\end{equation}
Likewise, we select actions that maximize both the current head's $Q$-values and the EVOI; \[arg\,max_a \, Q_h(s,a) +  EVOI(s,a)\]
We refer to this 
method
as \textit{BootDQN-EVOI} for convenience.

Our methods guide exploration 
in a more implicit manner compared to intrinsic 
motivation \cite{bellemare2016,burda2019exploration}, which directly alters rewards and affect loss computation. Instead of modifying the reward signal, our approach influences the action selection strategy, allowing 
the underlying algorithm (in this case \textit{BootDQN}) to assess action values. By increasing the desirability of certain actions, our methods encourage exploration that is likely to yield new insights. 
Notably, both $gain$ and EVOI are not simply state-specific, but also 
action-specific, which naturally allows for even more targeted exploration.

Both strategies are realized in 
Algorithm~2, 
where we get \textit{BootDQN-EVOI} when 
$aggregate=True$ and \textit{BootDQN-Gain}  
otherwise. Specifically, in 
line 1 of this algorithm, we perform a 
forward pass and receive a matrix 
of $Q$-values for each head and 
available action. In lines 2 and 3, we 
calculate the average $Q$-values per 
action over all heads and store the 
best and second-best actions according 
to the mean values. In lines 5-7, if 
$aggregate=True$, we compute the 
$gain_h$ for each head 
and then average over all heads 
to obtain EVOI. Finally, the action 
that maximizes the sum of active head's $Q$-values 
and EVOI is returned. On the other 
hand, if $aggregate=False$, we 
compute and utilize $gain_h$ for the active head $h$ and 
return the action that maximizes the 
sum of active head's $Q$-values and $gain_h$. 
It is important to stress that despite using Bootstrapped DQN 
as a basis for our algorithms, both techniques could be employed in any  methodology 
that involves multiple value-based  models.

\begin{figure}[t]
\raggedright
\hrule
\textbf{Algorithm 2} Action selection utilizing gain of information \\ 
\hrule
\renewcommand{\algorithmicrequire}{\textbf{Input:}}
\centering
\label{algorithm:gain}
\textbf{Input}: state $s$, $K$ value functions $Q_k(\cdot)$, active function $h$, boolean $aggregate$
\begin{algorithmic}[1]
    \STATE Obtain $Q_k(s,a)$ for each model $k \in [1,K]$ and action $a$
    \STATE Compute the average $Q$-values $\overline{Q}(s,a)$ over all models
    \STATE Compute best and second-best  (according to $\overline{Q}(s,a)$) actions $a^*_1$ and $a^*_2$
    \IF{$aggregate$}
        \STATE Compute $gain_k(s,a)$ $\forall a,k$ using Equation~\ref{equation:gain}
        \STATE Compute $EVOI(s,a) \forall a$ using Equation~\ref{equation:vpi}
        \RETURN{$arg\,max_{a} \, (Q_h(s,a) +  EVOI(s,a))$}
    \ELSE
        \STATE Compute $gain_h(s,a) ,\, \forall a$ using Equation~\ref{equation:gain}
        \RETURN{$arg\,max_{a} \, (Q_h(s,a) +  gain_h(s,a))$}
    \ENDIF
\end{algorithmic}
\hrule
\end{figure}

Having presented our methods,
we revisit our Fig.~\ref{fig:$EVOI$_showcase} example to analyze the effects of $gain$ and EVOI at a specific state in Pacman. 
Figure~\ref{fig:$EVOI$_showcase} displays the $Q$-values for each action alongside the enhancements provided by $gain$ and EVOI. 
At first 
glance, both techniques appear to work as a 
correction to the $Q$-values. $gain$ is 
sparser than EVOI, affecting only the LEFT 
action, as it is calculated from a single head's perspective. 
By contrast, EVOI, being the average $gain$ over 
all heads, shows denser but lower adjustments. 
In this scenario, 
{\em BootDQN} would choose 
DOWN greedily, while EVOI slightly boosts UP, 
promoting it as the selected action. $gain$, on the other hand,
significantly boosts LEFT.
Intuitively, the values of $gain$ reflect the 
extent of disagreement between the active head 
and the ensemble, while EVOI measures collective 
disagreement across all heads. 
For example, high $gain$ for the LEFT action 
indicates significant disagreement from the 
active head but EVOI is lower for the same action, which indicates that most heads  
seem to agree on its value. 
This illustrates how EVOI smooths outliers, 
providing more balanced estimates of expected 
information gain.

Now, a natural consideration for 
our methods is whether they may 
converge prematurely (due to 
them encouraging the network 
heads to resolve differences) 
and how well they manage different levels of  diversity. We investigate 
this issue (among others) 
in the following section, in which we detail the experimental evaluation of our algorithms.

\section{Experimental Evaluation}
\label{section:experiments}

In this section we present results from  experiments that test the performance 
of \textit{BootDQN-Gain} and \textit{BootDQN-EVOI} on selected hard-exploration Atari games. 
As a baseline, we also 
experiment with
our own implementation of
\textit{BootDQN} and \textit{BootDQN-UCB} that utilizes UCB~\cite{chen2017ucbexplorationqensembles}: 
\[arg\,max_a \, \overline{Q}(s,a) +  std(Q(s,a))\] for action selection, where $std(Q(s,a))$ is the empirical standard deviation of the ensemble. 
In Appendix A, 
we include additional results in 
a simpler 
environment called {\em DeepSea}~\cite{osband2019deep}.

\begin{figure*}[t]
    \centering
    \includegraphics[width=0.95\textwidth]{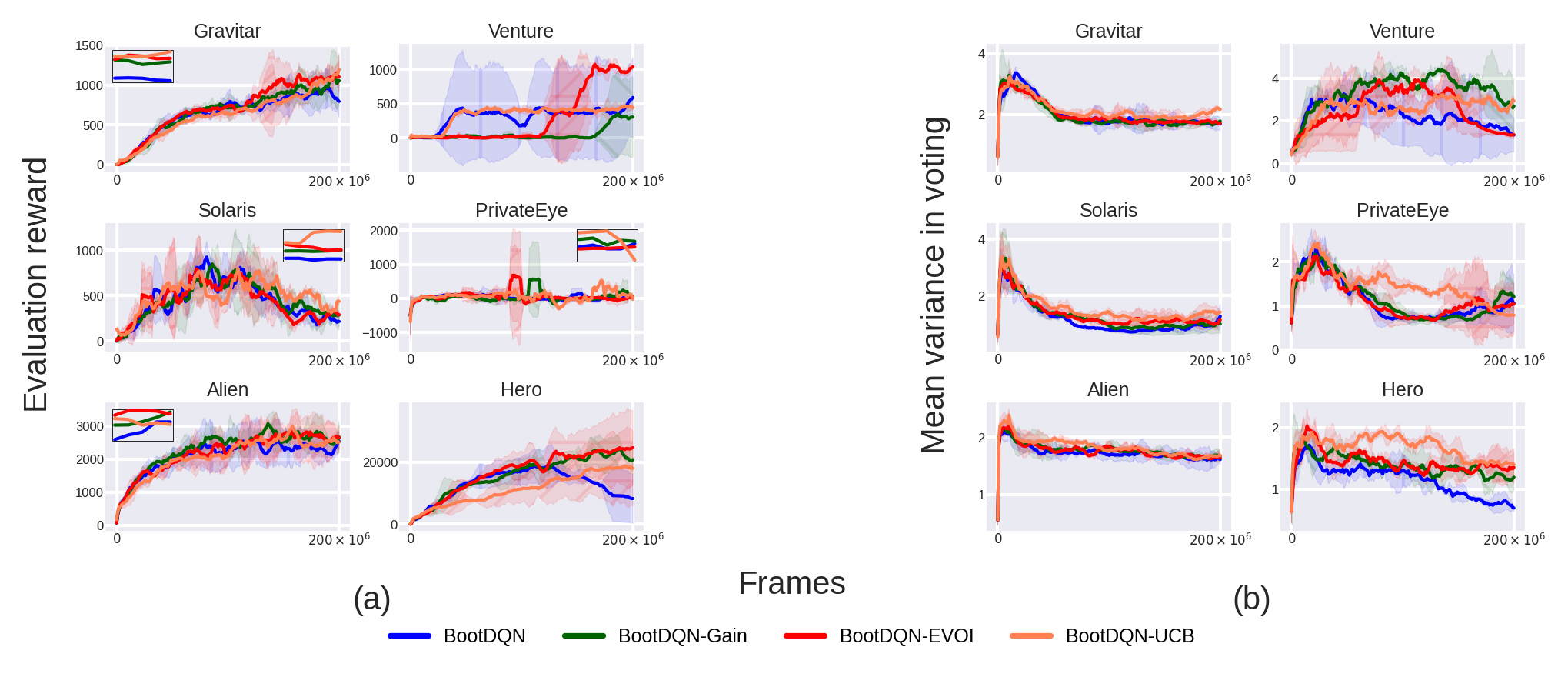}
    \caption{
     {\em Evaluation rewards} (subfigure (a)) and
    {\em mean variance in voting} (subfigure (b)) during evaluation periods,  
    for \textit{BootDQN} (our implementation), 
    \textit{BootDQN-UCB}, \textit{BootDQN-Gain},  
    and \textit{BootDQN-EVOI}. 
    All plots show moving averages using a sliding window of the 10 most recent values, with 
    shaded areas representing the 95\% confidence intervals for each algorithm. In terms of performance, \textit{BootDQN-EVOI} outperforms or matches its counterparts in all games. 
    For clarity, mini-plots 
    are included in cluttered plots in 
    (a) to magnify results for the last 
    5M frames.}
    \label{fig:diversity}
\end{figure*}

\subsection{Environment Preprocessing}

We apply some standard preprocessing 
techniques to the environment as 
introduced in \cite{mnih2015human}. 
We skip 4 frames of the environment for each chosen action,
by repeating each action for 4 consecutive frames. 
Each frame is converted from RGB to grayscale and rescaled to be of shape 
84$\times$84. The four most recent observations are stacked, creating a 4$\times$84$\times$84  
tensor that is used as an input to the network. 
To enhance stability and allow for a universal learning rate across all games, we also normalize each pixel's value so that it lies in $[0,1]$ 
and clip the rewards to $\{-1,0,+1\}$. Given the deterministic nature of Atari games, we introduce some stochasticity by 
forcing the agent to do nothing (no-operation) for a random number of timesteps (up to 30) at the start of each episode.

\subsection{Setting and Parameters}

Training is conducted for 50M steps, 
equivalent to 200M frames for a 4-frame skip 
per action. Each algorithm is run three times 
per game with different random seeds. 
The number of seeds is the same as in paper~\cite{nikolov2019information} 
that introduces \textit{DQN-IDS}, which we compare against. 
Hyperparameters are largely based on \cite{osband2016deep}, except for using the 
Adam optimizer \cite{KingBa15}, a different 
learning rate, and normalizing the total loss 
instead of the network gradients (see Section \ref{section:our_algorithms}). 
These changes were determined by preliminary tests 
on a few games. 
The parameters that we do not specify here, 
are assigned the PyTorch \cite{Ansel_PyTorch_2_Faster_2024} 
and Gymnasium \cite{towers2024gymnasium} 
default values.

To be comparable with existing results in the literature, all tested methods employ $K=10$ heads branching 
from a shared convolutional network, which 
consists of three convolutional layers: 
the first with 32 channels, 8$\times$8 kernel, 
and stride 4; the second with 64 channels, 4$\times$4 kernel, and stride 2; and the third 
with 64 channels, 3$\times$3 kernel, and 
stride 1. 
The output flows into each head, composed of 
two fully connected linear layers---one with 512 neurons and the 
final with neurons equal to the number of actions. 
The rectified linear unit activation is used 
after each layer, except for the last, which 
outputs the estimated $Q$-values. 
We set the the data sharing parameter equal to 
1, and use the Adam optimizer with a learning 
rate of $10^{-4}$. 
The replay buffer holds up to 1M experiences, 
with policy network updates every 4 steps and 
target network updates every 10K steps. 
Recall that each head has its own 
target network. 
The size of each mini-batch used for loss 
computation is set to 32. 
To reduce overestimation bias we consider 
Double DQN updates and to improve stability 
we utilize the Huber loss function. Training begins after collecting at least 50K 
transitions. The discount 
factor is set to 0.99. 

To measure performance, 
we 
run an evaluation process
every 1M frames. 
During evaluation, we pause 
network 
training
and run 10 episodes or 500K steps (whichever comes first) using 
majority voting between heads. 
Preliminary testing revealed that 
a short warm-up period allows for the heads 
to diversify and mitigates early convergence.

\subsection{Results}

Fig. \ref{fig:diversity}(a) 
presents the performance of
our algorithms: \textit{BootDQN}, \textit{BootDQN-UCB}, \textit{BootDQN-Gain} and \textit{BootDQN-EVOI}.

The games tested require increased exploration capabilities and 
are considered very demanding,   
as most of them 
offer few and sparse rewards, and  feature very long horizons 
compared to easier, more episodic,  
games; thus requiring 
deep exploration, as per the 
taxonomy presented in~\cite{bellemare2016}. 

Each plot represents 
the moving average of the algorithms' 
evaluation scores with a size $10$ sliding 
window for smoothing. It is evident 
that \textit{BootDQN-EVOI} outperforms its 
counterparts in \emph{Gravitar} and \emph{Venture} 
while in \emph{Solaris} we get similar performance 
although \textit{BootDQN-Gain} is a bit higher in 
the second half of training whereas 
\textit{BootDQN-EVOI} does slightly better in the first half. In \emph{Private Eye}, \textit{BootDQN-EVOI} and \textit{BootDQN-gain} 
learn a significantly better policy 
mid-learning, but cannot repeat this  
success consistently. 
\textit{BootDQN-UCB} 
demonstrates comparable 
performance overall, 
except for \emph{Hero}. 

\begin{table*}[t]
    \centering
    \caption{We present the maximal evaluation performance for BootDQN (our implementation), BootDQN-Gain and 
    BootDQN-EVOI. BootDQN$^*$ contains raw scores (maximal results) reported in  \cite{osband2016deep} and DQN-IDS results from  \cite{nikolov2019information}. For example, in \emph{Gravitar} BootDQN-Gain attained a 
    maximum evaluation score of 1663.3. Best scores in respective categories are styled in bold.}
    \begin{tabular}{lcccccc}
        \hline
         & \textit{BootDQN}$^*$ & \textit{BootDQN} & \textit{DQN-IDS} & 
        \textit{BootDQN-UCB} &
         \textit{BootDQN-Gain} & \textit{BootDQN-EVOI} \\
         \hline
         
         \emph{Gravitar} & 286.1 & 1480 & 771 & 1621.7  & 1663.3 & \textbf{1876.6} \\
         
         \emph{Solaris} & N/A & 2223.4 & 2086.8 & 2875.4 & 2252.4 & \textbf{3018.2}\\
         
         \emph{Private Eye} & 1812.5 & 1584.4 & 201.1 & \textbf{7811.1} & 5350.3 & 5880 \\
         
         \emph{Venture} & 212.5 & 1473.3 & 389.1 & 746.7 & 736.6 & \textbf{1543.3} \\
         
         \emph{Alien} & 2436.6 & 4291.3 & \textbf{9780.1} & 4040.7 & 5230.3 & 4364.6 \\
         
         \emph{Hero} & 21021.3 & 21537 & 15165.4 & 22453.2 & 28309 & \textbf{28539.3} \\
         \hline
    \end{tabular}
    
    \label{tab:results}
\end{table*}

In Table~\ref{tab:results} we report 
the 
returns of the best evaluation period 
throughout training. For our algorithms---that were run for 3 seeds, as in~\cite{nikolov2019information}---we take 
the maximum evaluation payoff for each 
seed and report the average. 
\textit{BootDQN}$^*$ and \textit{DQN-IDS} list the results reported 
in \cite{osband2016deep} and \cite{nikolov2019information}, 
respectively. On the other hand, 
\textit{BootDQN} represents our implementation 
of Bootstrapped DQN. 
Notably, our \textit{BootDQN} implementation of the 
algorithm attains higher scores than \textit{BootDQN}$^*$ (as implemented in the original Bootstrapped DQN paper). We attribute  
this to differences in the optimizer, learning rate, and our approach to loss normalization.

Overall, as seen in Table~\ref{tab:results}, \textit{BootDQN-EVOI} 
attains the best maximum performance in 
all games except for \emph{Alien} and \emph{Private Eye}. 
Interestingly, even when 
all algorithms have similar performance 
(as seen in Fig.~\ref{fig:diversity}(a)) 
\textit{BootDQN-EVOI} achieves much higher 
maximal scores. 
For example, 
in \emph{Solaris} \textit{BootDQN-EVOI} has performance 
akin to \textit{BootDQN}, but the former succeeds 
a much higher maximal evaluation period, 
2223.4 versus 3018.2. This 
 means that \textit{BootDQN-EVOI} 
attains---albeit momentarily---high 
returns, which however it cannot repeat long enough 
 to effectively learn them.

Furthermore, to aggregate the 
performance of all methods, we compute 
the human-normalized 
score (HNS) as in \cite{van2016deep}: \[normalized\_score = \frac{agent\_score - random\_score}{human\_score - random\_score}\]for the maximal reward of each game and present the mean scores over all games 
in Table \ref{tab:norm_results}. 
\begin{table}[t]
    \centering
        \caption{Mean Human-Normalized Score (HNS) for each algorithm over all tested games,  computed from the maximal evaluation scores reported in Table~\ref{tab:results}. BootDQN-EVOI has 9.38\% increased performance compared to the baseline.}
    \setlength{\tabcolsep}{1mm}
    \begin{tabular}{lc}
        \hline
         & Mean HNS \\
         \hline
         \textit{BootDQN} & $50.67$\% \\
         \textit{DQN-IDS} & $40.89$\% \\
         \textit{BootDQN-UCB} & $43.6$\% \\
         \textit{BootDQN-Gain} & $48.3$\% \\
         \textit{BootDQN-EVOI} & $\bf{60.05}$\% \\
         \hline
    \end{tabular}
    \label{tab:norm_results}
\end{table}
\textit{BootDQN-EVOI} has the best performance overall in 
these hard testbeds, outperforming the baseline by 9.38\%, \textit{DQN-IDS} by 19.16\%, and \textit{BootDQN-UCB} by 16.45\%. Human and 
random scores were taken from  
\cite{nikolov2019information}.

Finally, we measure the uncertainty among heads by calculating the mean variance 
of action-votes during each evaluation period, which is depicted in Fig.~\ref{fig:diversity}(b). We also compute the mean variance by averaging over all evaluation periods and report the results in Table~\ref{tab:variance}. 
In general, 
we found it common for \textit{BootDQN} to exhibit low initial diversity that gradually increases to a peak value before decreasing steadily or plateauing. 
Despite this trend, \textit{BootDQN-Gain} 
and \textit{BootDQN-EVOI} maintain higher 
variance (Fig.~\ref{fig:diversity}(b), Table~\ref{tab:variance}) than \textit{BootDQN} in most games, either during significant periods of training or in the final variance levels. 
Note that \textit{BootDQN-UCB} explicitly maintains even higher  
variance most of the time, but ultimately achieves lower performance than \textit{BootDQN-EVOI}. 
We believe that the increased variance likely arises from selecting controversial 
actions that can lead to further controversial 
situations. Notably, although our algorithms 
encourage the heads to learn similar $Q$-values, two heads can have nearly 
identical values, and consequently low EVOI/$gain$, yet still vote for 
different actions. This is possible since majority 
voting does not account for differences 
in value magnitude.

\begin{table}[t]
    \setlength{\tabcolsep}{0.98mm}
    \centering
    \caption{Mean variance during evaluation voting, across all evaluation periods. We observe that both BootDQN-EVOI and BootDQN-Gain maintain mean variance equal to or greater than the baseline, while BootDQN-UCB exhibits the highest mean variance across all games, except for \emph{Venture}.}
    \begin{tabular}{lcccc}
        \hline
        & \textit{BootDQN} & \textit{BootDQN-UCB} & \textit{BootDQN-Gain} & \textit{BootDQN-EVOI} \\
        \hline
        \emph{Gravitar} & 2 & 2.15 & 1.96 & 2 \\
        \emph{Venture} & 2.24 & 2.62 & 3.15 & 2.52 \\
        \emph{Solaris} & 1.14 & 1.52 & 1.25 & 1.32 \\
        \emph{Private Eye} & 1.12 & 1.43 & 1.16 & 1.12 \\
        \emph{Alien} & 1.72 & 1.82 & 1.78 & 1.76 \\
        \emph{Hero} & 1.14 & 1.67 & 1.38 & 1.43 \\
        \hline
    \end{tabular}
    
    \label{tab:variance}
\end{table}

\subsection{Performance with Different Levels of Diversity}

We now turn our attention to 
studying the impact of increased diversity of the ensemble model to learning performance.
This is an interesting problem, since the diversity of ensemble models has been identified in the literature as a key factor for the superior performance demonstrated by ensemble models in many settings~\cite{ortega2022diversity}. 
Trivially, in \textit{BootDQN}, 
increasing $K$ 
also
increases the diversity of the 
overall model, so the easiest way to 
fluctuate diversity is to adjust 
$K$. 

\begin{figure}[t]
    \centering
\includegraphics[width=0.8\columnwidth]    {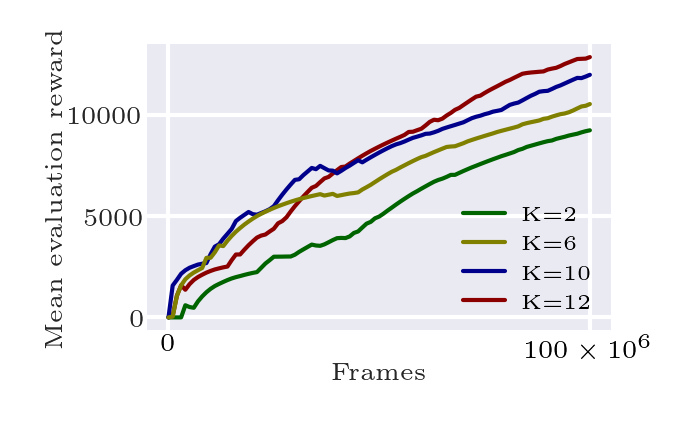}
    \caption{Mean evaluation reward of \textit{BootDQN-EVOI} as training progresses, for different number of heads $K$ in \emph{Hero}. Performance improves as $K$ increases.}
    \label{fig:diversity_analysis}
\end{figure}

As \textit{BootDQN-EVOI} was the best-performing method in our previous experiments, we focus on studying its behavior in the face of increased levels of diversity. 
In Fig.~\ref{fig:diversity_analysis} 
we demonstrate the performance of \textit{BootDQN-EVOI} 
in the Atari game of \emph{Hero} as we increase the 
number of heads $K$. 
Specifically, we test for 
$K \in \{2, 6, 10, 12\}$ and 
observe that a higher $K$ translates to 
better overall performance. 
Thus, there is no indication for premature convergence and, by increasing the number of heads, performance improves. 

Of course, any decision to increase the number of heads to improve performance should be taken while considering time scalability concerns and related computational constraints.
Regarding time scalability, Table~\ref{tab:rumtime} presents \textit{BootDQN-EVOI}'s runtime for different values of $K$ for {\em Hero} on a machine with an AMD Ryzen 7 3700x CPU, an NVIDIA RTX 3080 GPU and 32GB of RAM.
We also note that for $K=10$  \textit{BootDQN-gain} exhibits similar runtimes to  \textit{BootDQN-EVOI}, while  \textit{BootDQN} is approximately 10\% faster.

\begin{table}[t]
    \centering
    \caption{Runtimes for different number of heads $K$. These correspond to Fig.~\ref{fig:diversity_analysis}, and represent training time of BootDQN-EVOI for 100M frames on \emph{Hero}. We list the rounded increase in percentage compared to the previous $K$. For example, for $K=6$, training required 
    36\% more wall time than for $K=2$.}
    \label{tab:rumtime}
    \begin{tabular}{lcc}
        \hline
            & Runtime & \% difference to previous $K$\\
            \hline
            $K$=2 & 1d 53m & -- \\
            $K$=6 & 1d 9h 53m & +36\% \\
            $K$=10 & 1d 15h 3m & +15\% \\
            $K$=12 & 1d 18h 59m & +10\% \\
         \hline
    \end{tabular}
\end{table}

\section{Conclusions and Future Work}
\label{section:conclusions}

In this paper we studied the problem of enhancing exploration in 
ensemble DRL by utilizing 
the notion of
action-related 
{\em information gain}. To this end, we put forward two closely-related algorithms we developed, 
\textit{BootDQN-Gain} and \textit{BootDQN-EVOI}.
Both of our algorithms
measure the expected benefit of exploratory actions 
through discrepancies in $Q$-values. This approach 
is different from existing work in this direction, as we estimate the potential of each action in providing new information, rather than selecting actions with values that neural network models are uncertain about. 
Our experiments demonstrate that 
\textit{BootDQN-EVOI} outperforms 
Bootstrapped DQN 
(and {\em DQN-IDS}) in 
performance on several hard Atari games. 
In conclusion, the integration of EVOI in ensemble 
DRL presents a promising direction for advancing 
exploration strategies and improving learning performance. 

In terms of future work, we remark that our methods' impact to the 
algorithm's learning is delayed, since 
the produced data samples by the enhanced 
action selection are stored in the 
replay buffer and may not be used for an 
update after many steps. 
Using a smaller 
replay buffer could be a naive way to 
overcome this, and would ultimately hinder  performance. We suggest that 
our methods would benefit from incorporating a 
more efficient sampling method, such as 
proportional prioritized experience 
replay~\cite{van2016deep}. 
Finally, we anticipate that 
employing {\em voting 
mechanisms} as alternatives to the simple majority voting used by {\em BootDQN} and variants, during both training and evaluation, will lead to even better 
utilization of the  knowledge gathered. 

\section*{Acknowledgments}

The research described in this paper was carried out within the framework of
the National Recovery and Resilience Plan Greece 2.0, funded by the European Union - NextGenerationEU (Implementation Body: HFRI. Project name:
DEEP-REBAYES. HFRI Project Number 15430).

\bibliographystyle{IEEEtran}
\bibliography{references}

\appendices
\label{appendix}

\section{Additional Experimental Results in DeepSea}

Here we present the results of further 
experimentation in the \emph{DeepSea}~\cite{osband2019deep, osband2018randomized}---a tabular, grid-based, 
environment that tests the deep-exploration 
capabilities of algorithms. 

\begin{figure}[!h]
    \centering
\includegraphics[width=0.8\columnwidth]    {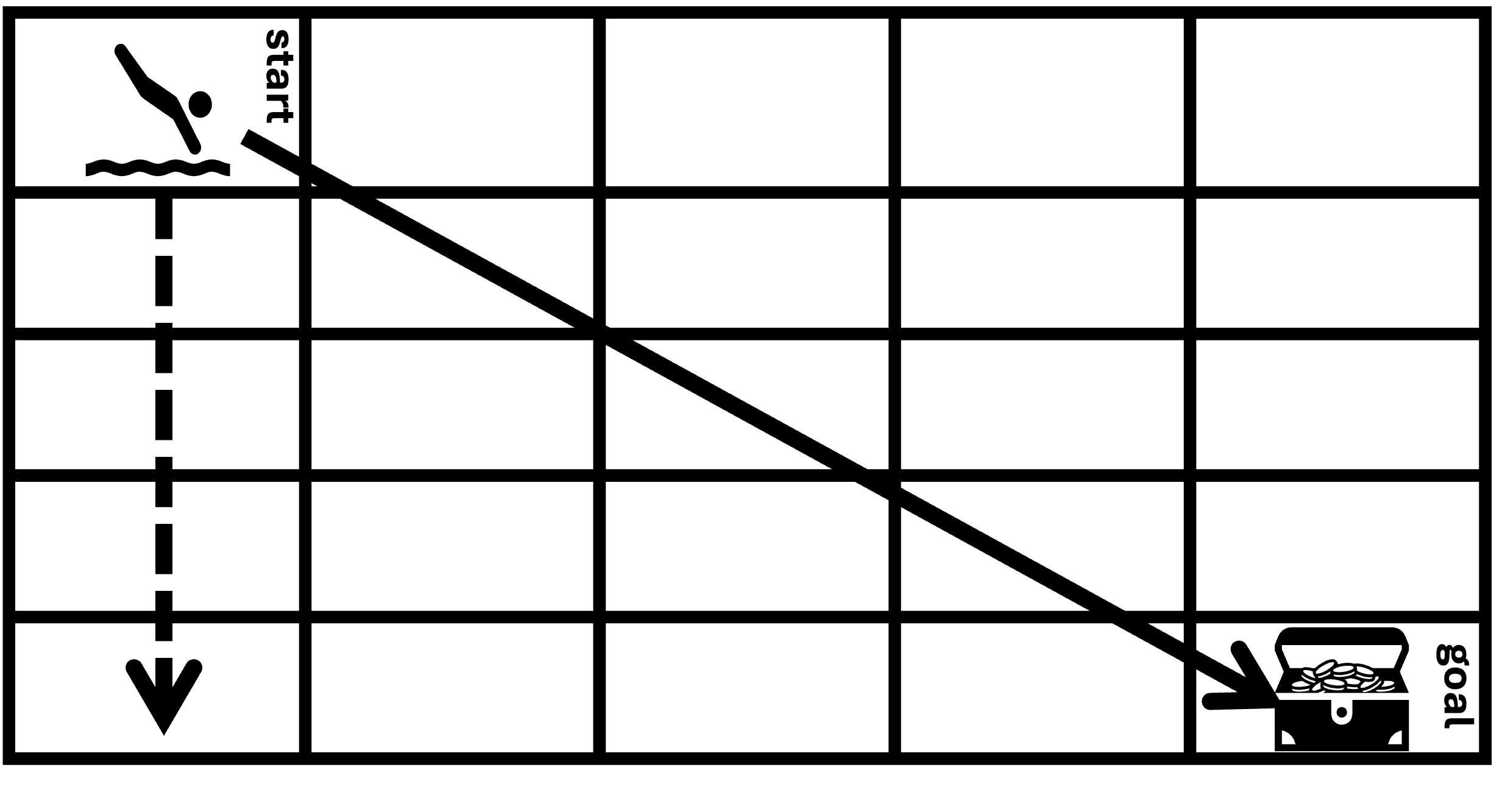}
    \caption{Example instance of a DeepSea problem of size $N$ ($N$x$N$ grid). The agent starts each episode at the top leftmost cell and at each timestep descends one row and can either move left or right. The goal is to reach the bottom rightmost cell and perform the action right. The agent receives a small negative reward for choosing right and a substantially positive reward for reaching the goal. Solid arrow represents the optimal policy ($\sum_tr_t=0.99$) and dashed arrow represents the second-best -- suboptimal -- policy ($\sum_tr_t=0$).}
    \label{fig:deepsea_example}
\end{figure}

\begin{table}[!h]
    \centering
    \caption{Hyperparameters used for DeepSea experiments}
    \begin{tabular}{lc}
        Max. number of episodes & 100k\\
        Number of networks & 20\\
        Optimizer & Adam\\
        Learning rate & 0.001\\
        Replay buffer length & 10k\\
        Batch size & 128\\
        Data sharing (Bernoulli parameter) & 0.5\\
        Target network update frequency & $N$ (size of grid)\\
        Number of hidden layers & 2\\
        Hidden layer size & 50\\
    \end{tabular}
    \label{tab:deepsea_hyperparams}
\end{table}

An example of this environment is illustrated in Fig.~\ref{fig:deepsea_example}. Note that we can increase the difficulty of the problem by increasing 
the dimensions of the $N \times N$ grid. 
As we do that, it becomes progressively 
harder and an algorithm would need to explore 
persistently in order to discover the optimal solution and not get misled by the deceptive feedback. 
This happens because the environment yields negative rewards 
when the agent moves right and no reward when the agent 
moves left, but the optimal solution can be achieved only 
by selecting right at each one of the $N$ timesteps.

Each algorithm was trained $15$ times for different initial seeds, using the hyperparameters in Table~\ref{tab:deepsea_hyperparams}. 
These hyperparameters were largely based on~\cite{osband2018randomized}.
Each run was terminated when either {\em (a)} the algorithm's 
policy converges to the optimal one; or {\em (b)} a predetermined number of episodes is reached. 
Also, for \textit{BootDQN-EVOI} specifically, we use 
$EVOI(s,a) = \sum_{k \in K}gain_k(s,a)$. 
We assume that an agent has converged to the optimal solution when the running average regret drops below $0.9$.

\begin{figure}[!h]
    \centering
\includegraphics[width=1\columnwidth]    {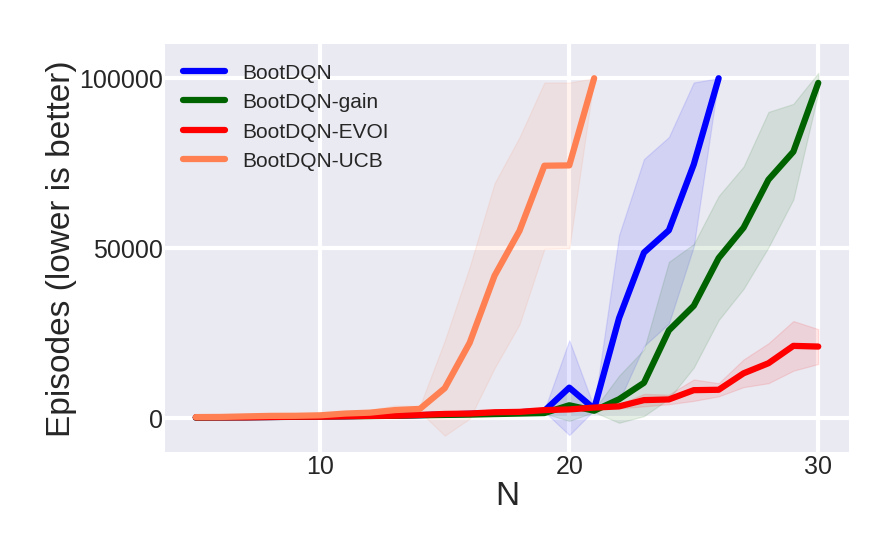}
    \caption{Average number of episodes needed, for each algorithm, to solve the corresponding DeepSea instance over $15$ seeds. Shaded areas visualize the 95\% confidence intervals. Both of our methods outperform the baselines. \textit{BootDQN-EVOI} in particular scales the best as the size of the problem (i.e., the $N$ of the $N \times N$ grid) increases. }
    \label{fig:deepsea}
\end{figure}

The results are presented in Fig.~\ref{fig:deepsea}, where we plot the average number 
of episodes needed (lower is better) for each one of the tested methods for $N$ ranging from $5$ to $30$. Evidently, \textit{BootDQN-EVOI} is the method that scales best 
as $N$ increases. 

\end{document}